\documentclass[runningheads]{llncs}

\usepackage{amsmath,amsfonts,bm}

\def\eqref#1{equation~\ref{#1}}

\def\1{\bm{1}}

\def\vb{{\bm{b}}}

\def\vg{{\bm{g}}}
\def\vh{{\bm{h}}}

\def\vx{{\bm{x}}}
\def\vy{{\bm{y}}}

\def\evb{{b}}

\def\evg{{g}}
\def\evh{{h}}

\def\mW{{\bm{W}}}

\DeclareMathAlphabet{\mathsfit}{\encodingdefault}{\sfdefault}{m}{sl}
\SetMathAlphabet{\mathsfit}{bold}{\encodingdefault}{\sfdefault}{bx}{n}

\def\sL{{\mathbb{L}}}

\def\sN{{\mathbb{N}}}

\def\sX{{\mathbb{X}}}

\usepackage{hyperref}
\usepackage{url}
\usepackage{booktabs}
\usepackage{array}
\usepackage{optidef}
\usepackage{xcolor}
\usepackage{tabularx}
\usepackage{amsthm}
\usepackage{placeins}
\usepackage{lscape}

\usepackage{subcaption}
\usepackage{alphalph}

\newcommand{\moons}{\texttt{MOONS}\,}

\usepackage{graphicx}
\usepackage{epigraph} 

\begin{document}
\title{Training Thinner and Deeper Neural Networks: Jumpstart Regularization
}
%
%
\author{Carles Riera\inst{1}
\and 
Camilo Rey\inst{1}
\and
Thiago Serra\inst{2}
\and
Eloi Puertas\inst{1}
\and 
Oriol Pujol\inst{1}}
\authorrunning{C. Riera et al.}
%
\institute{Universitat de Barcelona, Spain\\
\email{crieramo8@alumnes.ub.edu} \\
\email{camilorey@gmail.com} \\
\email{\{epuertas,oriol\_pujol\}@ub.edu}
\and
Bucknell University, United States\\
\email{thiago.serra@bucknell.edu}}
\maketitle              
\begin{abstract}
Neural networks are more expressive when they have multiple layers. 
In turn, conventional training methods are only successful if 
the depth does not lead to numerical issues such as exploding or vanishing gradients, 
which occur less frequently when the layers are sufficiently wide. 
However, increasing width to attain greater depth entails the use of heavier computational resources and leads to overparameterized models. 
These subsequent issues have been partially addressed by model compression methods such as quantization and pruning, some of which relying on normalization-based regularization of the loss function to make the effect of most parameters negligible.
In this work, 
we propose instead to use regularization for preventing neurons from dying or becoming linear, a technique which we denote as \emph{jumpstart regularization}.  
In comparison to conventional training, 
we obtain neural networks that are thinner, deeper , and---most importantly---more parameter-efficient.


\end{abstract}

\section{Introduction}

\epigraph{\textit{Leap, and the net will appear.}}{Anonymous}

Artificial neural networks are inspired by the simple, yet powerful idea that predictive models can be produced by combining units that mimic biological neurons. 
In fact, there is a rich discussion on what should constitute each unit and how the units should interact with one another. 
Units that work in parallel form a layer, whereas a sequence of layers transforming data unidirectionally define a feedforward network. 
Deciding the number of such layers---the \emph{depth} of the network---is yet a topic of debate and technical challenges.

A neural network is trained for a particular task by minimizing the loss function associated with a sample of data in order for the network to learn a function of interest. 
Although several universal approximation results show that mathematical functions can generally be approximated to arbitrary precision by single-layer feedforward networks, 
these results rely on using a very large number of units \cite{kolmogorovnn,universalsigmoid,universalrelu}. 
Moreover, 
simple functions such as XOR cannot be exactly represented with a single layer using the most typical units \cite{minsky69perceptrons}.

In fact, 
it is commonly agreed that depth is important in neural networks \cite{bengio2014representation,lecun2015DeepLearningBig}.
In the popular case of feedforward networks in which each unit is a Rectified Linear Unit~(ReLU)~\cite{OriginReLU,nair2010rectified,ReLUGood2,lecun2015DeepLearningBig}, 
the neural network models a piecewise linear function~\cite{arora2018understanding}. 
Under the right conditions, the number of such ``pieces''---the \emph{linear regions}---may grow exponentially on the depth of the network~\cite{pascanu2013on,montufar2014on,telgarsky2015representation}. 
Depending on the total number of units and size of the input, the number of linear regions is maximized with more or less layers~\cite{serra2018bounding}. 
Similarly, there is an active area of study on bounding the number of layers necessary to model any function that a given type of network can represent~\cite{minsky69perceptrons,eldan2016power,arora2018understanding,gribonval2020approximation,vardi2021size,hertrich2021lower}.

Although shallow networks present competitive accuracy results in some cases~\cite{ba2014do}, 
deep neural networks have been established as the state-of-the-art over and again in areas such as computer vision and natural language processing~\cite{lenet,lstm,alexnet,vgg16,He2015BigIntoRELU,huang2017densely,transformers,bert} 
thanks to the %
the development and popularization of backpropagation~\cite{Werbos:81sensitivity,rumelhart1986learning,lecun1988theoretical}. 
However, Stochastic Gradient Descent~(SGD)~\cite{StochasticApproximation}---the training algorithm associated with backpropagation---may have difficulties to converge to a good model due to exploding or vanishing gradients \cite{vanishing1,vanishing2,pascanu2013difficulty,BengioSimardFrasconi94}.

Exploding and vanishing gradients are often attributed to excessive depth, inadequate choice of parameters for the learning algorithm, or inappropriate scaling between network parameters, inputs, and outputs \cite{Glorot10Initialization,batchnorm}. 
This issue has also inspired unit augmentations~\cite{leaky,prelu,crelu}, additional connections across layers~\cite{resnet,huang2017densely}, and output normalization~\cite{batchnorm,farkas}. 
Indeed, it is somewhat intuitive that gradient updates, depth, and parameter scaling may affect one another. 

In lieu of reducing depth, we may also increase the number of neurons per layer~\cite{wideresnet,inceptionv1,efficientnet,efficientnetv2,simpnet}. 
That leads to models that are considerably more complex, 
and which are often trained with additional terms in the loss function such as weight normalization 
to induce simpler models that hopefully generalize better. 
In turn, that helps model compression techniques such as network pruning methods to remove several parameters with only minor impact to model accuracy.

Nonetheless, vanishing gradients may also be caused by  \emph{dead} neurons when using ReLUs. 
If dead, a ReLU only outputs zero for every sample input. Hence, it does not contribute to updates during training and neither to the expressiveness of the model. 
To a lesser but relevant extent, similar issues can be observed with a RELU which never outputs zero, which we refer to as a \emph{linear} neuron.  

In this work, 
we aim to reverse neurons which die or become linear during training. 
Our approach is based on satisfying certain constraints throughout the process. For a margin defined for each unit, at least one input from the sample is above and another input is below. 
For each layer and input from the sample, at least one unit in the layer has that input above such a margin and another unit has it below. 
In order to use SGD for training, these constraints are dualized as part of the loss function and thus become a form of regularization that would prevent converging with the original loss function to spurious local minima.

\section{Background}

We consider a feedforward neural network modeling a function $\hat{\vy} = f_\theta(\vx)$ with an input layer $\vx = \vh^0 = [\evh_1^0 ~ \evh_2^0 \dots \evh_{n_0}^l]^T$, $L$ hidden layers, and each layer $\ell \in \sL = \{1,2,\dots,L\}$ having $n_\ell$ units indexed by $i \in \sN_\ell = \{1, 2, \ldots, n_\ell\}$. 
For each layer $\ell \in \sL$,
let $\mW^\ell$ be the $n_\ell \times n_{\ell-1}$ matrix in which the $j$-th row corresponds to the weights of neuron $j$ in layer $\ell$ and  $\vb^\ell$ be vector of biases of layer $\ell$. 
The preactivation output of unit $j$ in layer $\ell$ is $\evg_i^\ell = \mW_{j}^\ell \vh^{\ell-1} + \evb_j^\ell$ and the output is $\evh_j^\ell = \sigma(\evg_j^\ell)$ for an activation function $\sigma$, 
which if not nonlinear would allow hidden layer $\ell$ to be removed by directly connecting layers $\ell-1$ and $\ell+1$ \cite{serra2020lossless}. 
We refer to $\vg^\ell(\chi)$ and $\vh^\ell(\chi)$ as the values of $\vg^\ell$ and $\vh^\ell$ when $\vx = \chi$.

For the scope of this work, 
we consider the ReLU activation function $\sigma(u) = \max\{0, u\}$. 
Typically, the output of a feedforward neural network is produced by a softmax layer following the last hidden layer~\cite{softmax}, 
$\hat{\vy} = \rho(\vh^L)$ with $\rho(\vh^L)_j = e^{\evh^L_j}/\sum_{k=1}^{n_{L}} e^{\evh^L_k} ~ \forall j \in \{1, \ldots, n_{L}\}$,
which is a peripheral aspect to our study.

The neural network is trained by minimizing a  loss function $\mathcal{L}$ over a parameter set $\theta := \{ (\mW^\ell, \vb^\ell) \}_{\ell=1}^L$ based on the $N$ samples of a training set $\sX := \{ (\vx^i) \}_{i=1}^N$ to yield predictions $\{ \hat{\vy}^i := f_\theta(\vx^i) \}_{i=1}^N$ that approximate the sample labels $\{ \vy^i \}_{i=1}^N$ using metrics such as least squares or cross entropy \cite{Goodfellow-et-al-2016,understandingMLshalev}:
\begin{align}
\min_{\theta} ~~~ & \mathcal{L}\left( \theta, \left\{ (\hat{\vy}^i, \vy^i) \right\}_{i=1}^N \right) & \label{eq:min} \\
\text{s.t.} ~~~ & \hat{\vy^i} = f_\theta(\vx^i) \qquad \qquad \forall i \in \{1, 2, \ldots, N\} \label{eq:yhat} 
\end{align}
Whereas a neural network is not typically trained through constrained optimization, we believe that our approach is more easily understood under such a mindset, 
which aligns with further work emerging from  this community~\cite{bienstock2018principled,icarte2019training,fischetti2019embedded}. 

\section{Death, Stagnation, and Jumpstarting}

Every ReLU is either \emph{inactive} if $\evg^\ell_i \leq 0$ and thus $\evh^\ell_i = 0$ 
or \emph{active} if $\evg^\ell_i > 0$ and thus $\evh^\ell_i = \evg^\ell_i > 0$. 
If a ReLU does not alternate between those states for different inputs, then the unit is considered \emph{stable}~\cite{tjeng2019evaluating} and thus the neural network models a less expressive function~\cite{serra2020empirical}. In certain cases, those units can be merged or removed without affecting the model~\cite{serra2020lossless,serra2021scaling}. 
We consider in this work a superset of such units---those which do not change of state at least for the training set:

\begin{definition}
For a training set $\sX$, 
unit $j$ in layer $\ell$ is dead if $\evh^\ell_j(\vx^i) = 0 ~ \forall i \in \{1, 2, \ldots, N\}$, linear if $\evh^\ell_j(\vx^i) > 0 ~ \forall i \in \{1, 2, \ldots, N\}$, or nonlinear otherwise. 
Layer $\ell$ dead or linear if all of its units are dead or linear, respectively. 
\end{definition}

\begin{figure}[t!]
\centering
\begin{subfigure}[t]{0.30\textwidth}
    \includegraphics[width=0.7\textwidth]{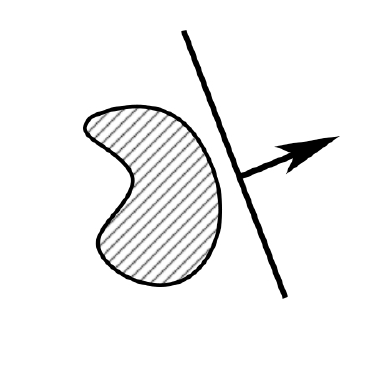}
    \caption{Dead unit}
    \label{fig:dead_unit}
    
\end{subfigure}
\begin{subfigure}[t]{0.30\textwidth}
    \includegraphics[width=0.7\textwidth]{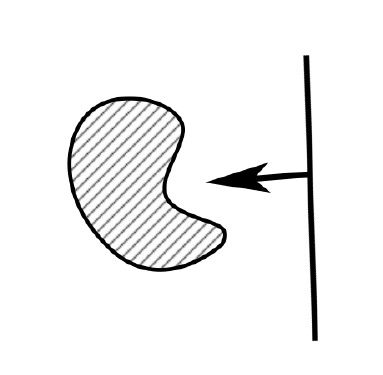}
    \caption{Linear unit}
    \label{fig:linear_unit}
    
\end{subfigure}
\begin{subfigure}[t]{0.30\textwidth}
    \includegraphics[width=0.7\textwidth]{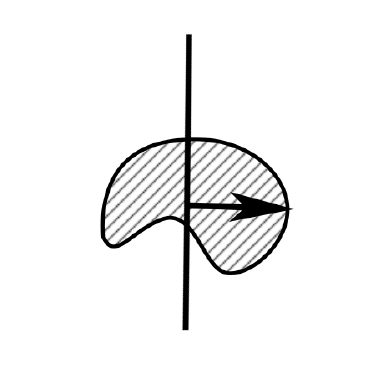}
    \caption{Nonlinear unit}
    \label{fig:non-linear_unit}
    
\end{subfigure}
\centering
\begin{subfigure}[t]{0.30\textwidth}
    \includegraphics[width=0.6\textwidth]{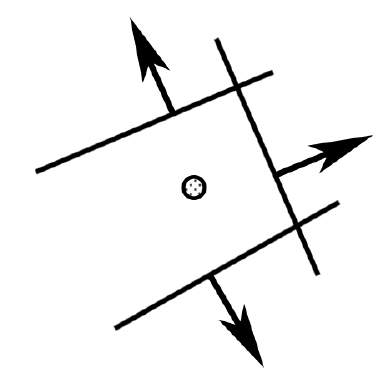}
    \caption{Dead point}
    \label{fig:dead_point}
    
\end{subfigure}
\begin{subfigure}[t]{0.30\textwidth}
    \includegraphics[width=0.6\textwidth]{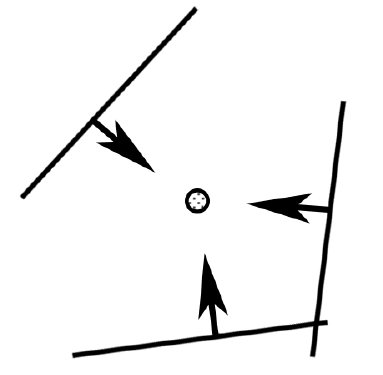}
    \caption{Linear point}
    \label{fig:linear_point}
    
\end{subfigure}
\begin{subfigure}[t]{0.30\textwidth}
    \includegraphics[width=0.6\textwidth]{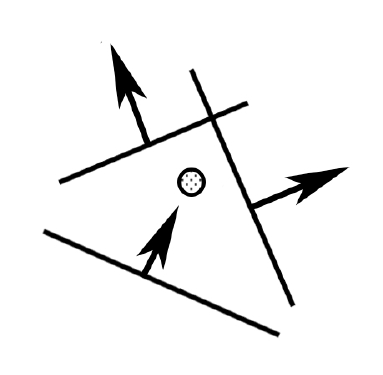}
    \caption{Nonlinear point}
    \label{fig:non-linear_point}
    
\end{subfigure}
\caption{A unit $j$ in layer $\ell$ separates the input space $\vh^{\ell-1}$ into an open half-space $\mW^\ell_j \vh^{\ell-1} + \evb^\ell_j > 0$ in which the unit is active and a closed half-space $\mW^\ell_j \vh^{\ell-1} + \evb^\ell_j \leq 0$ in which the unit is inactive. The arrow in each case points to the active side. 
The unit is dead if the inputs from training set $\sX$ lie exclusively on the inactive side (\ref{fig:dead_unit}); linear if exclusively on the active side (\ref{fig:linear_unit}); 
and nonlinear otherwise (\ref{fig:non-linear_unit}). 
In turn, an input is considered a dead point if it is in the closed half-space $\mW^\ell_j \vh^{\ell-1} + \evb^\ell_j\leq 0$ in which each and every unit $j \in \sN_\ell$ is inactive (\ref{fig:dead_point}); a linear point if it is in the open half-space $\mW^\ell_j \vh^{\ell-1} + \evb^\ell_j > 0$ in which each and every unit $j \in \sN_\ell$ is active (\ref{fig:linear_point}); and a nonlinear point otherwise (\ref{fig:non-linear_point}).}\label{fig:dead_linear_unit}
\end{figure}

Figures~\ref{fig:dead_unit} to \ref{fig:non-linear_unit} illustrate geometrically the classification of the unit based on the training set.  
If dead, a unit impairs the training of the neural network because it always outputs zero for the inputs in the training set. Unless the units preceding a dead unit are updated in such a way that the unit is no longer dead, then the gradients of its output remain at zero and the parameters of the dead unit are no longer updated~\cite{deadunit,deadUnit2}, 
which effectively reduces the modeling capacity. 
If a layer dies, then the training stops because the gradients are zero.

For an intuitive and training-independent discussion, we consider incidence of dead layers at random. 
If the probability that a unit is dead upon initialization is $p$, as reasoned in \cite{deadunit}, 
then layer $\ell$ is dead with probability $p^{n_\ell}$ and at least one layer is dead with probability $1 - \prod_{\ell=1}^L (1-p)^{n_\ell}$. 
If a layer is too thin or the network is too deep,  then the network is more likely to be untrainable. 
We may discard dead unit initializations, but that ignores the impact on the training set:

\begin{definition}
For a hidden layer $\ell \in \sL$,
an input $x$ is considered a dead point if $\vh^\ell(x) = 0$, a linear point if $\vh^\ell(x) > 0$, and a nonlinear point otherwise. 
\end{definition}

Figures~\ref{fig:dead_point} to \ref{fig:non-linear_point} illustrate geometrically the classification of a point based on the activated units. 
If $x^i \in \sX$ is a dead point at layer $\ell$, 
then there is no backpropagation associated with $x^i$ to the hidden layers $1$ to $\ell-1$. Hence, its contribution to training is diminished unless a subsequent gradient update at a preceding unit reverts the death. 
If $\ell = L$, then $x^i$  is effectively not part of the training set. 
If all points die, regardless of the layer, then training halts. 

If we also associate a probability $q$ for $\vx^i$ not activating a unit,  
then $\vx^i$ is dead for layer $\ell$ with probability $q^{n_\ell}$ and for at least one layer of the neural network with probability $1-\prod_{\ell=1}^L (1-q)^{n_\ell}$. 
Unlike $p$, $q$ is bound to be significant.

We may likewise regard linear units and linear points as less desirable than nonlinear units and nonlinear points. 
A linear unit limits the expressiveness of the model, 
since it always contributes the same linear transformation to every input in the training set. 
A linear point can be more difficult to discriminate from other inputs, in particular if those inputs are also linear points.

Inspired by the prior discussion, 
we formulate the following constraints: 
\begin{align}
& \max_{\vx^i \in \sX} g^\ell_j(\vx^i) \geq 1 & \forall \ell \in \sL, j \in \sN_\ell \label{eq:no_dead_unit} \\
& \min_{\vx^i \in \sX} g^\ell_j(\vx^i) \leq -1 & \forall \ell \in \sL, j \in \sN_\ell \label{eq:no_linear_unit}  \\ 
& \max_{j \in \sN_\ell} \evg^\ell_j(\vx^i) \geq 1 & \forall \ell \in \sL, \vx^i \in \sX \label{eq:no_dead_point} \\
& \min_{j \in \sN_\ell} \evg^\ell_j(\vx^i) \leq -1 & \forall \ell \in \sL, \vx^i \in \sX \label{eq:no_linear_point}
\end{align}
Dead and linear units are respectively prevented by the constraints in (\ref{eq:no_dead_unit}) and (\ref{eq:no_linear_unit}).
Dead and linear points are prevented by the constraints in (\ref{eq:no_dead_point}) and (\ref{eq:no_linear_point}). 
Then we dualize those constraints 
and 
induce their satisfaction through the objective:

\begin{align}
\min_{\theta} ~~~ & \mathcal{L}\left( \theta, \left\{ (\hat{\vy}^i, \vy^i) \right\}_{i=1}^N \right) + \lambda \mathcal{P}(\xi^+, \xi^-, \psi^+, \psi^-) \\
\text{s.t.} ~~~ & \hat{\vy^i} = f_\theta(\vx^i) & \forall i \in \{1, 2, \ldots, N\} \\
& \xi^+_{j \ell} = \max \left\{0, 1 - \max_{\vx^i \in \sX} g^\ell_j(\vx^i) \right\} & \forall \ell \in \sL, j \in \sN_\ell \label{eq:rel_dead_unit} \\
& \xi^-_{j \ell} = \max \left\{0, -1 - \min_{\vx^i \in \sX} g^\ell_j(\vx^i) \right\} & \forall \ell \in \sL, j \in \sN_\ell \label{eq:rel_linear_unit} \\ 
& \psi^+_{i \ell} = \max \left\{0, 1-\max_{j \in \sN_\ell} \evg^\ell_j(\vx^i) \right\} & \forall \ell \in \sL, \vx^i \in \sX \label{eq:rel_dead_point} \\
& \psi^-_{i \ell} = \max \left\{0, -1 - \min_{j \in \sN_\ell} \evg^\ell_j(\vx^i) \right\} & \forall \ell \in \sL, \vx^i \in \sX \label{eq:rel_linear_point}
\end{align}
We denote by $\xi^+$, $\xi^-$, $\psi^+$, and $\psi^-$ the nonnegative deficits associated with the corresponding constraints in (\ref{eq:no_dead_unit})--(\ref{eq:no_linear_point}) which are not satisfied. 
These deficits are combined and weighted against the original loss function $\mathcal{L}$ through a function $\mathcal{P}$, 
for which we have considered the arithmetic mean as well as the 1 and 2-norms.

We can apply this to convolutional neural networks \cite{neocognitron,lenet} with only minor changes, 
since they are equivalent to a feedforward neural network with parameter sharing and which is not fully connected. 
The main difference to work with them directly is that the preactivation of the unit is a matrix instead of a scalar. 
We compute the 
margin through 
the maximum or minimum over those values.

\section{Computational Experiments}

\begin{figure}[ht]
    \centering
    \begin{subfigure}[t]{0.45\textwidth}
        \includegraphics[width=\textwidth]{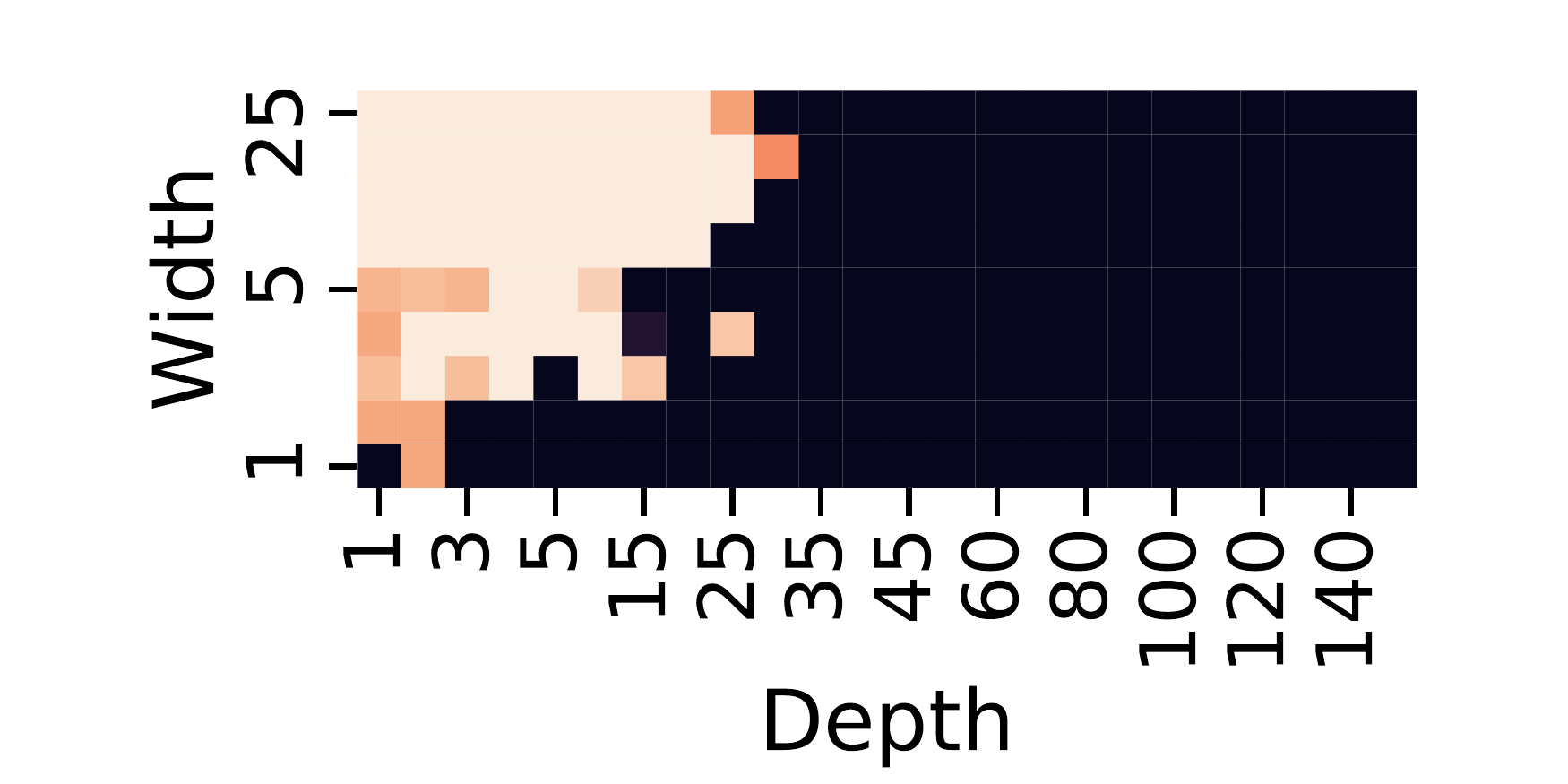}
        \caption{Train accuracy with baseline}
        \label{fig:moons_grid_relu_val}
    \end{subfigure}
    \centering
    \begin{subfigure}[t]{0.45\textwidth}
        \includegraphics[width=\textwidth]{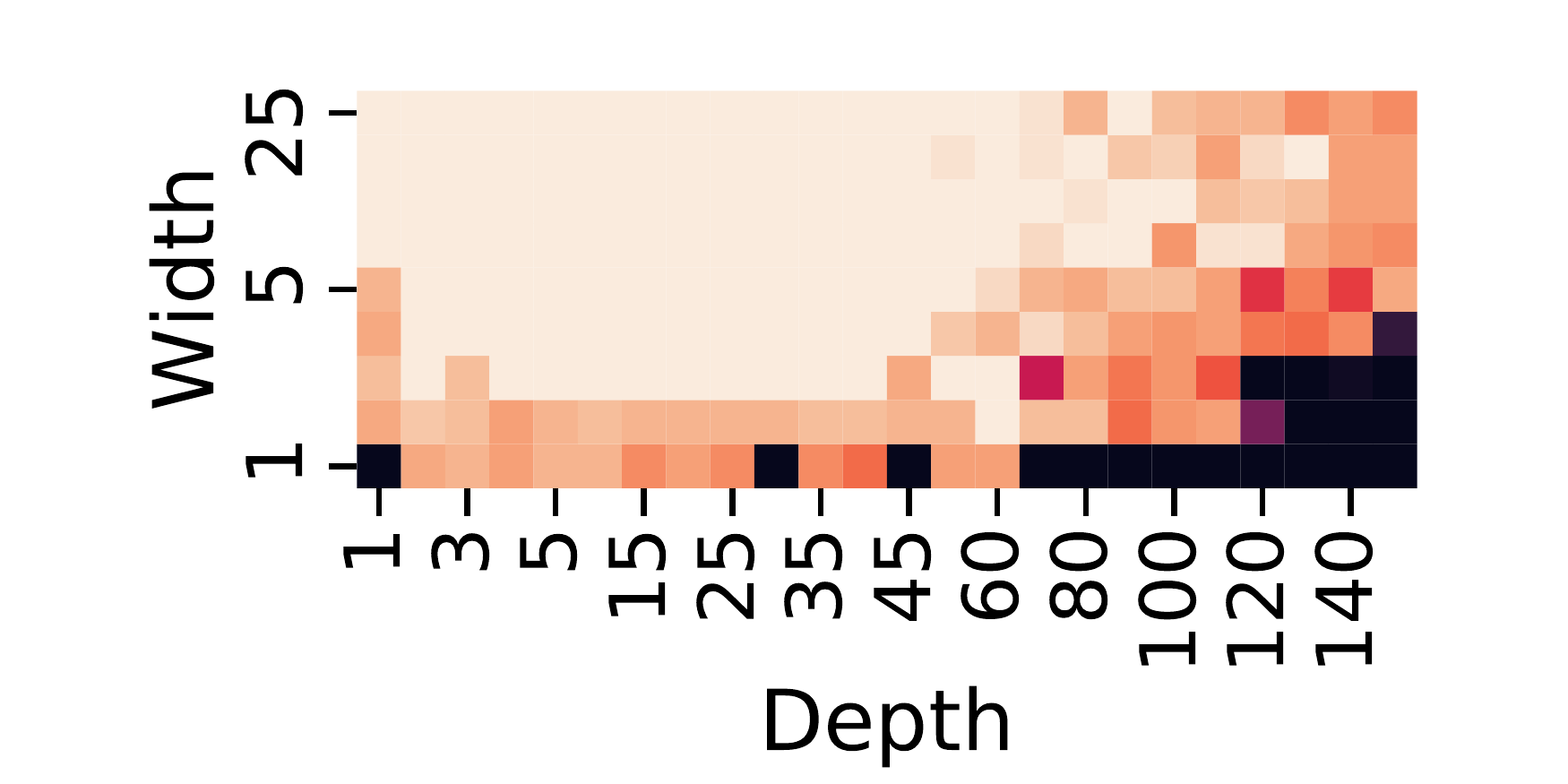}
        \caption{Train accuracy with jumpstart}
        \label{fig:moons_grid_up_val_00001}
    \end{subfigure}
    \centering
    \begin{subfigure}[t]{0.08\textwidth}
        \includegraphics[width=0.9\textwidth]{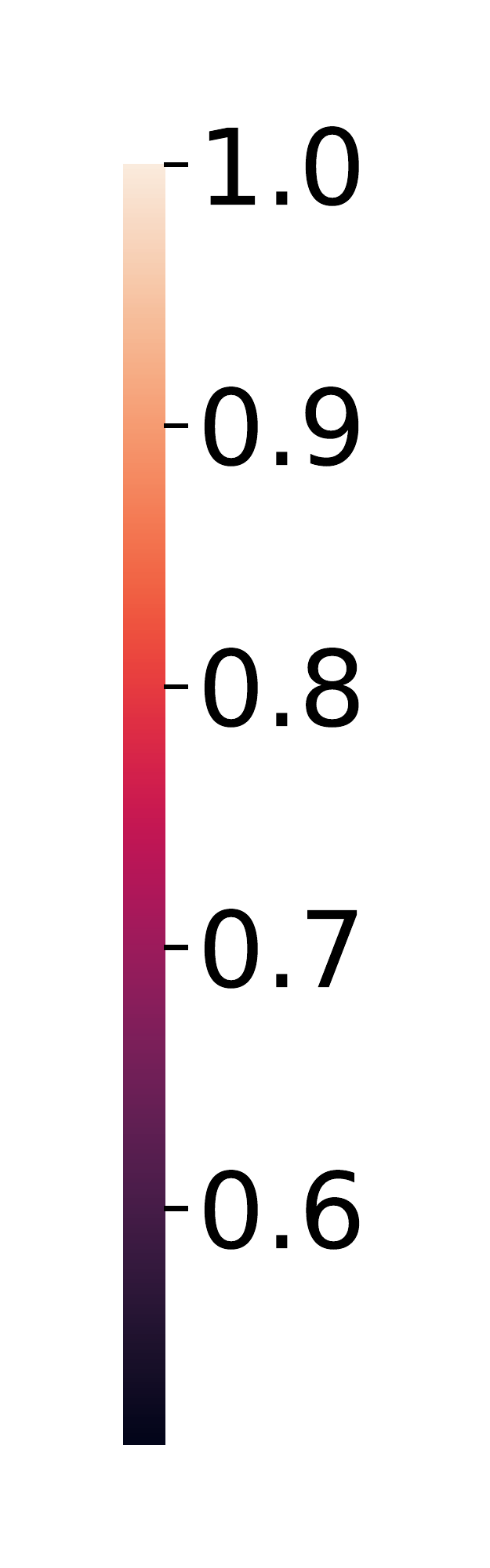}
    \end{subfigure}
    \centering
    \begin{subfigure}[t]{0.45\textwidth}
        \includegraphics[width=\textwidth]{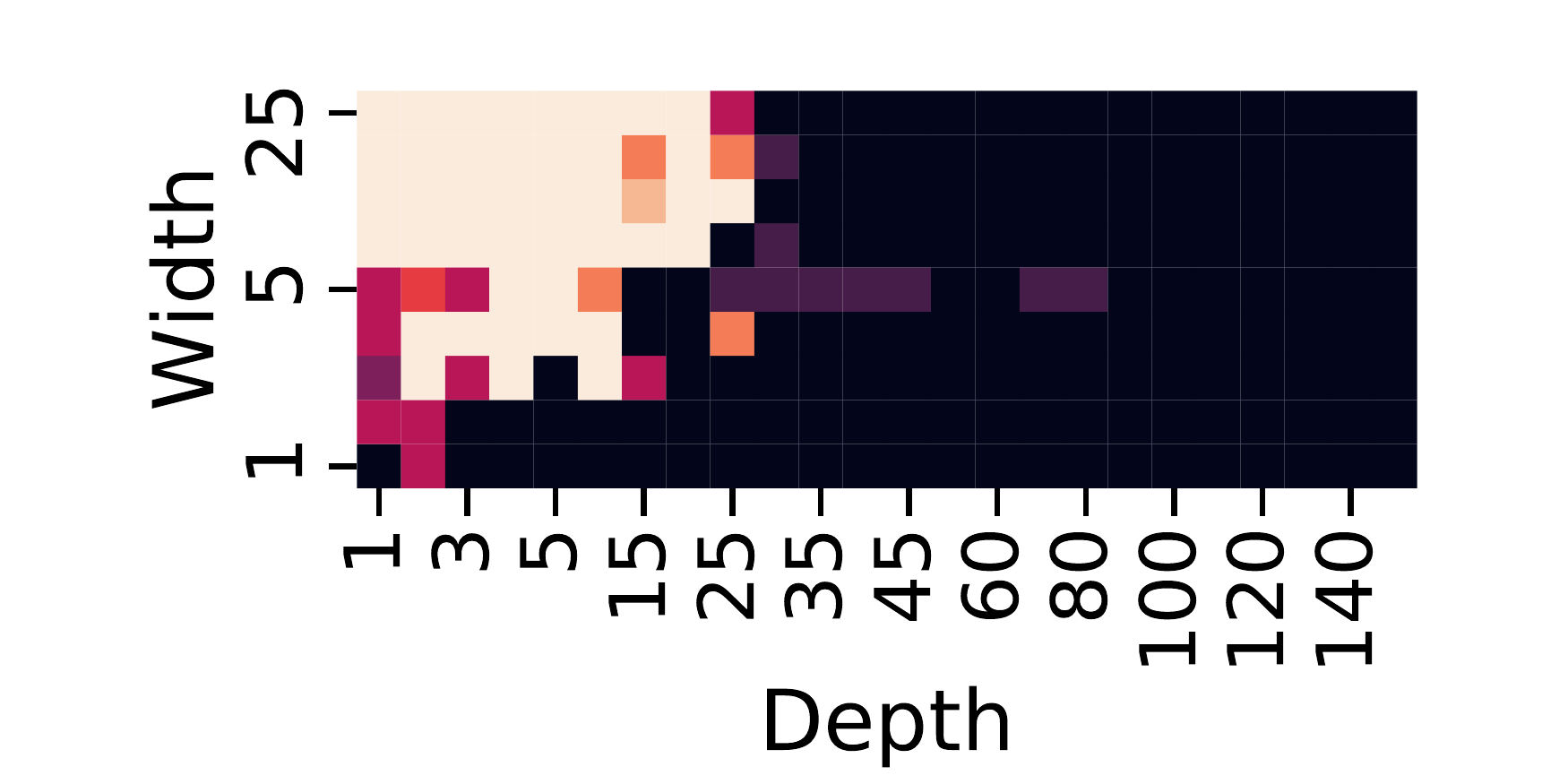}
        \caption{Validation accuracy with  baseline}
        \label{fig:moons_grid_relu_val}
    \end{subfigure}
    \centering
    \begin{subfigure}[t]{0.45\textwidth}
        \includegraphics[width=\textwidth]{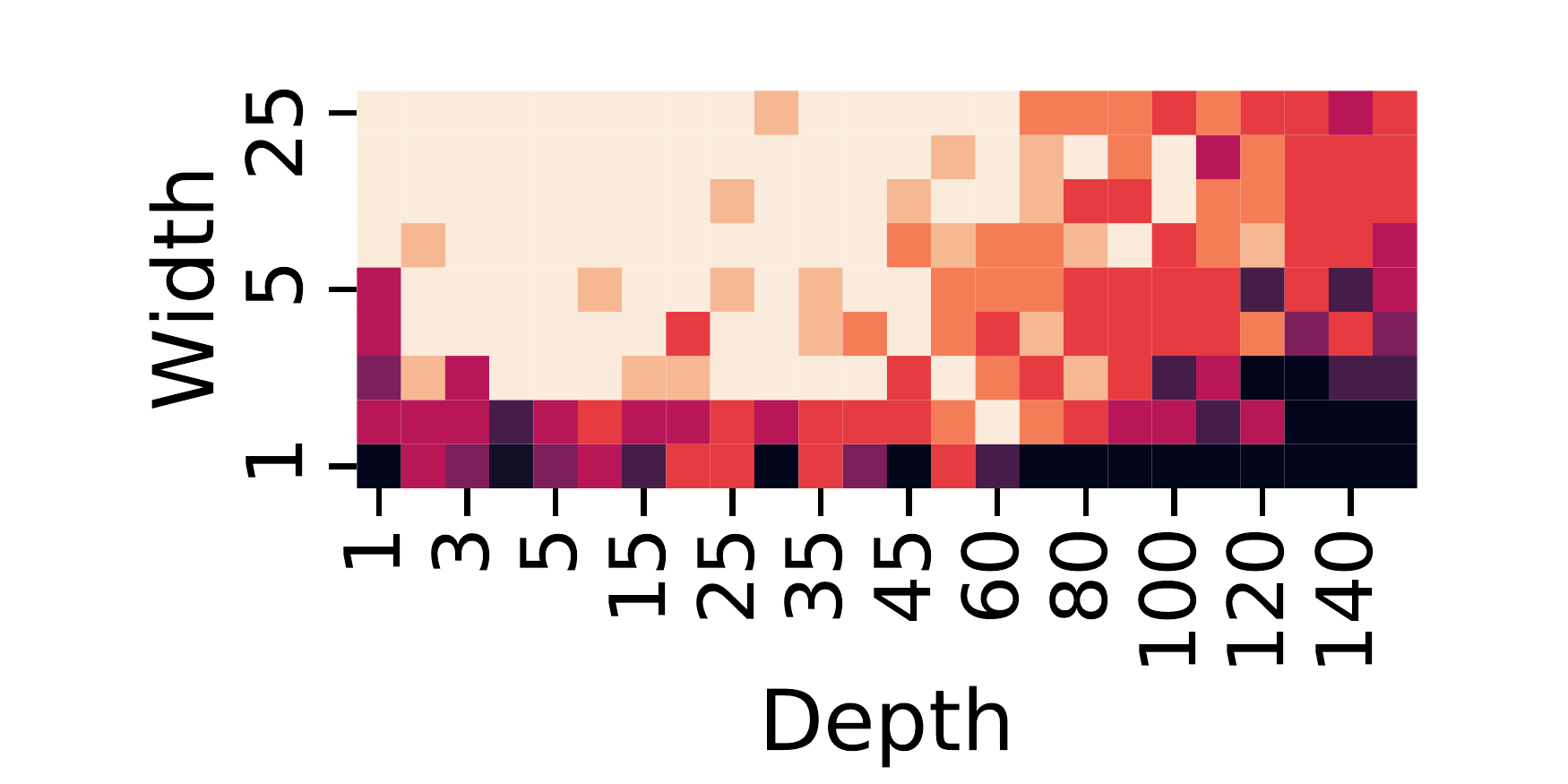}
        \caption{Validation accuracy with jumpstart}
        \label{fig:moons_grid_up_val_00001}
    \end{subfigure}
    \centering
    \begin{subfigure}[t]{0.08\textwidth}
        \includegraphics[width=0.9\textwidth]{images/moons_grid/cbar_lambda..0.0_mode..nan_negative_margin..-1.0-val_acc.pdf}
    \end{subfigure}

  \caption{Heatmap contrasting accuracy for neural networks trained on \moons with depth between 1 and 150 and width between 1 and 25. The left plot is the baseline and the right plot shows the results when using jumpstart. The accuracy ranges from a low of 0.5 (black) to a high of 1.0 (beige), with the former  corresponding to random guessing since the dataset has two balanced classes.}
  \label{fig:moons_grid} 
\end{figure}

Our first experiment (Figure \ref{fig:moons_grid}) is based on the \moons dataset \cite{scikit-learn} with 85 points for training and 15 for validation. We test every width in $\{1, 2, 3, 4, 5, 10, 15, 20, 25\}$ with every depth in $\{1, 2, 3, 4, 5, 10, 15, 20,  25, 30, 35, 40, 45, 50, 60, \ldots, 150 \}$. 
We chose a simpler dataset to limit the inference of factors such as overfitting, underfitting, or batch size issues. The  networks are implemented in \texttt{Tensorflow} \cite{tensorflow} and \texttt{Keras} \cite{keras} with Glorot uniform initialization \cite{Glorot10Initialization} and trained using Adam \cite{adam} for $5000$ epochs,  learning rate of $\epsilon = 0.01$,  and  batch size of $85$.
For each depth-width pair, 
we train a baseline network and a network with jumpstart using 1-norm as the aggregation function $\mathcal{P}$ and loss coefficient 
$\lambda = 10^{-4}$. 

With jumpstart, 
we successfully train networks of width 3 with a depth up to 60 instead of 10 for the baseline 
and width 25 with a depth of up to 100 instead of 30. 
Hence, there is an approximately 5-fold increase in trainable depth. 

\begin{table}[t]
\caption{Summary of the results for the convolutional neural networks trained on the MNIST dataset without jumpstart (baseline) and with jumpstart.}
\label{tab:mnist}
\centering
\begin{tabular}{@{\extracolsep{6pt}}ccccc}
\noalign{\vskip1.5pt} 
    & \multicolumn{2}{c}{Baseline} & \multicolumn{2}{c}{Jumpstart} \\
    \noalign{\vskip1.5pt} 
    \cline{2-3} \cline{4-5}
    \noalign{\vskip1.5pt} 
     & Training & Validation & Training & Validation \\
     \noalign{\vskip1.5pt} 
     \cline{2-2}
     \cline{3-3}
     \cline{4-4}
     \cline{5-5}
     \noalign{\vskip1.5pt} 
     Best overall accuracy & 0.999467 & 0.9885 & 0.999533 & 0.9911 \\
     Successful model & 18 & 18 & 54 & 54 \\
     Best for depth-width pair & 8 & 11 & 45 & 41 \\
\end{tabular}
\end{table}

Our second experiment (Table \ref{tab:mnist}) evaluates convolutional neural networks trained on the MNIST dataset \cite{mnist}. 
We test every depth from $2$ to $68$ in increments of $4$ with every width in $\{ 2, 4, 8 \}$, 
where the width refer to the number of filters per layer. 
The networks are implemented as before, but with a learning rate of $0.001$ over $50$ epochs, batch size of $1024$, 
kernel dimensions $(3,3)$, padding to produce an output of same dimensions as the input, Glorot uniform initialization \cite{Glorot10Initialization}, flattening before the output layer and using a baseline and a jumpstart network with 1-norm as the aggregation function $\mathcal{P}$ and loss coefficient $\lambda = 10^{-8}$. 

With jumpstart, 
we successfully train networks combining all widths and depths in comparison to only up to depth 12 for widths 2 and 4 and only up to depth 24 for width 8 in the baseline. 
In other words, only 18 baseline network trainings converge, which we denote as the successful models in Table \ref{tab:mnist}. 

\begin{figure}[t]
\begin{center}
    \begin{subfigure}[t]{1\textwidth}
        \includegraphics[width=\textwidth]{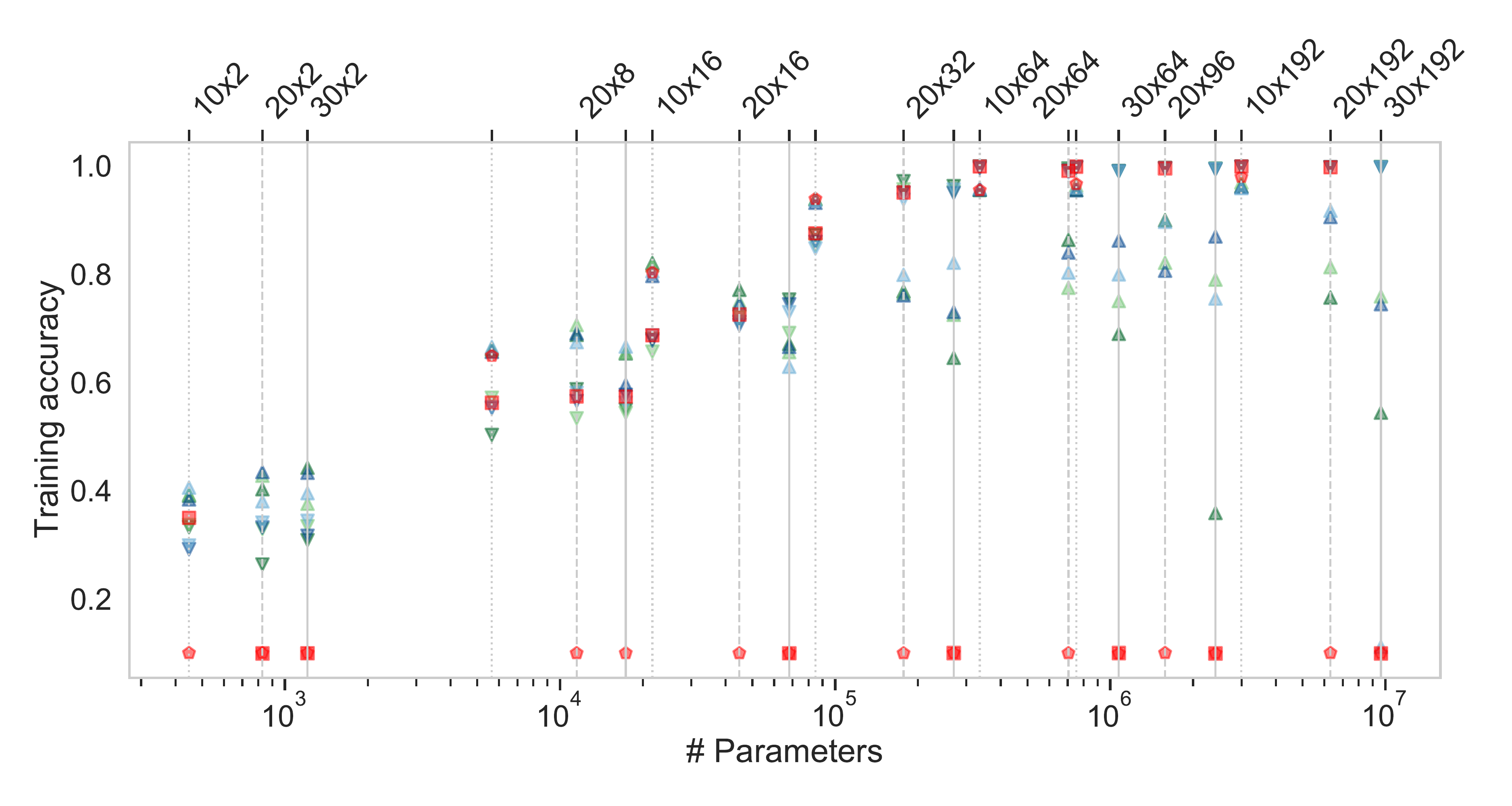}

    \end{subfigure}
    ~
   \begin{subfigure}[t]{1\textwidth}
        \includegraphics[width=\textwidth]{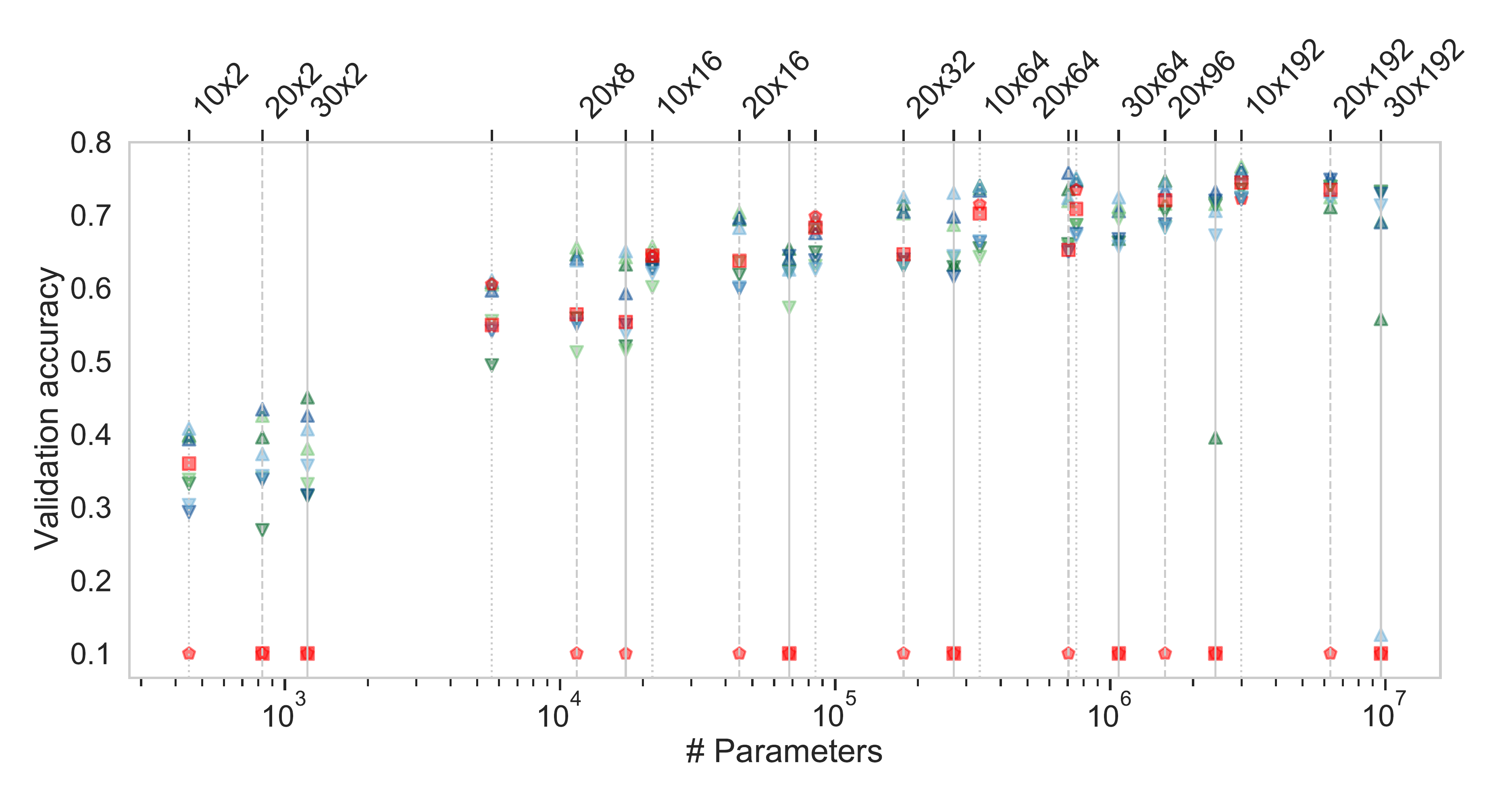}

    \end{subfigure}
    ~
    \begin{subfigure}[t]{0.75\textwidth}
        \includegraphics[width=\textwidth]{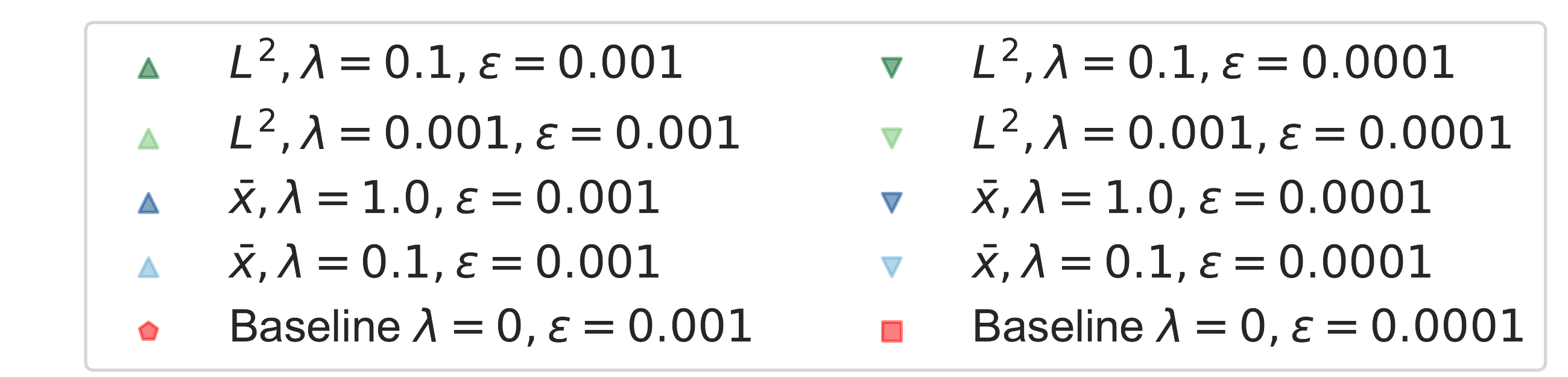}
    \end{subfigure}
    ~
\end{center}
\caption{Scatter chart of the number of parameters by accuracy for training (top) and validation (bottom) of convolutional neural networks trained on CIFAR-10. Some depth-width pairs are shown above the plots for reference and the gridlines are solid for depth $30$, dashed for $20$, and dotted for $10$. The results of this experiment are plotted in this format due to their greater variability in comparison to the second experiment, which permits evaluating parameter efficiency. With same number of units but fewer parameters, the results for $20 \times 8$ are better than $10 \times 16$ and likewise for $20 \times 32$ when compared with $10 \times 64$.}
\label{fig:cifar_param_count}
\end{figure}

\begin{figure}[t]
\begin{center}
    \begin{subfigure}[t]{1\textwidth}
        \includegraphics[width=\textwidth]{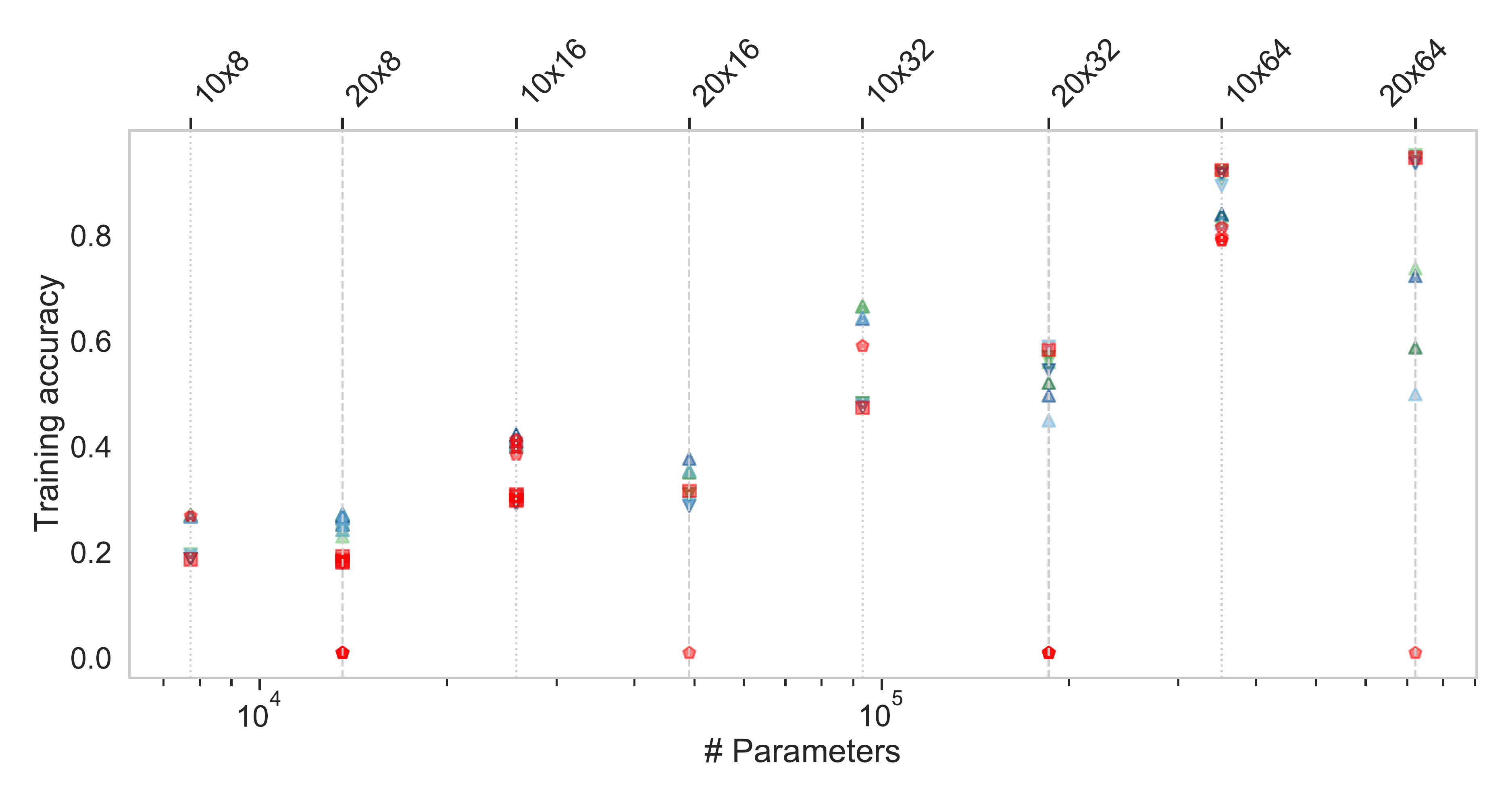}

    \end{subfigure}
    ~
   \begin{subfigure}[t]{1\textwidth}
        \includegraphics[width=\textwidth]{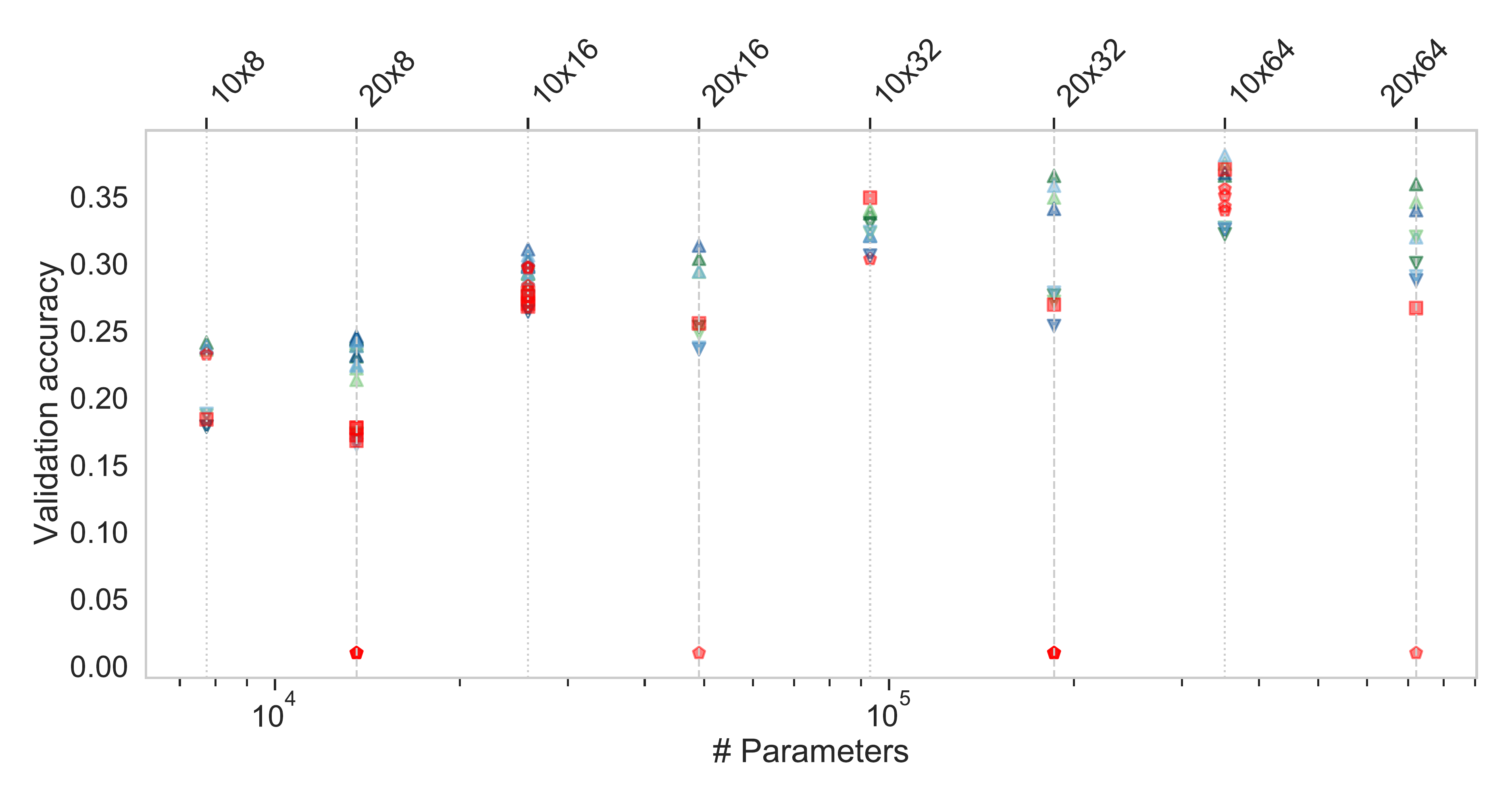}

    \end{subfigure}
    ~
\end{center}
\caption{Scatter chart of number of parameters by accuracy for training (top) and validation (bottom) of convolutional neural networks trained on CIFAR-100. Some depth-width pairs are shown above the plots for reference and the gridlines are dashed for depth $20$ and dotted for $10$. Once certain capacity is reached at $640$ units, we find that the performance for $20 \times 32$ is competitive with that of $10 \times 64$ while using less parameters.
}
\label{fig:cifar100_param_count}
\end{figure}

Our third experiment (Figures \ref{fig:cifar_param_count} and \ref{fig:cifar100_param_count}) evaluates convolutional networks trained on CIFAR-10 and CIFAR-100 \cite{cifar10}. 
For CIFAR-10, we test every depth in $\{ 10, 20, 30 \}$ with every width in $\{ 2, 8, 16, 32, 64, 96, 192 \}$. 
For CIFAR-100, 
we test depths in $\{ 10, 20 \}$ with widths in $\{ 8, 16, 32, 64 \}$. 
The networks are implemented in Pytorch \cite{pytorch}, with learning rates $\varepsilon \in \{0.001, 0.0001\}$ over $400$ epochs, batch size of $128$, 
same kernel dimensions and padding, Kaiming uniform initialization \cite{He2015BigIntoRELU}, global max-avg concat pooling before the output layer, and jumpstart with 2-norm ($\mathcal{P} = L^2$) and $\lambda \in \{0.001, 0.1\}$ or mean ($\mathcal{P} = \bar{x}$) 
and $\lambda \in \{0.1, 1\}$. 

With jumpstart, 
we successfully train networks for CIFAR-10 with depth up to 30 in comparison to no more than 20 in the baseline. 
The best performance---$0.766$ for jumpstart and $0.734$ for baseline---is observed for both with $\varepsilon = 0.001$, 
where the validation accuracy of each jumpstart experiment exceeds the baseline in  18 out of 21 depth-width pairs in one case and 20 out of 21 in another.
The baseline is comparatively more competitive with $\varepsilon = 0.0001$, but the overall validation accuracy drops significantly. 
For CIFAR-100, the jumpstart experiments exceed the baseline in 12 out of 16 combinations of depth, width, and learning rate. The accuracy improves by $1$ point in networks with $10$ layers and $7.8$ points in networks with 20 layers. 
The maximum accuracy attained is $0.37$ for the baseline and $0.38$ with jumpstart. 
The training time becomes $1.33$ times greater in CIFAR-10 and $1.47$ in CIFAR-100. 
The use of the precomputed pre-activations on the forward pass involves a similar memory cost: around 50\% more.

The source code is at \url{https://github.com/blauigris/jumpstart-cpaior}.

\section{Conclusion}

We have presented a regularization technique for training thinner and deeper neural networks, 
which leads to a more efficient use of the dataset and to neural networks that are more parameter-efficient. 
Although massive models are currently widely popular in theory \cite{ntk} and practice  \cite{amodei2018aicompute}, 
their associated economical barriers and environmental footprint \cite{strubell2019policy} as well as societal impact \cite{parrots} are known concerns. 
Hence, we present a potential alternative to lines of work such as model compression  \cite{blalock2020survey} by avoiding to operate with larger models. 
Whereas deeper networks are often pursued, trainable thinner networks are surprisingly not.

\FloatBarrier
%
%
%

\section*{Acknowledgements ~~}
Thiago Serra was supported by the National Science Foundation (NSF) grant IIS 2104583.

\bibliographystyle{splncs04}
\bibliography{references}

\end{document}